%% file: main.tex
\documentclass[runningheads]{llncs}
\usepackage[T1]{fontenc}
\usepackage{graphicx}

\usepackage{subcaption}
\usepackage{xcolor}
\usepackage{multirow}
\usepackage{algorithm}
\usepackage{float}
\usepackage{algpseudocode}

\begin{document}
%
%\title{Training Deep Learning Architectures in synthetic procedurally-generated forest environments and validate them in real-life cases\thanks{Supported by organization x.}}
%\title{Use of synthetic procedurally-generated forest environments in point-based deep learning architectures \thanks{Supported by organization x.}}
\title{Training point-based deep learning networks for forest segmentation with synthetic data}
\titlerunning{Point-based deep learning for forest segmentation}
\author{Francisco {Raverta Capua}\inst{1}\orcidID{0009-0004-3337-7741} \and
Juan {Schandin}\inst{2}\orcidID{0009-0009-9994-3503} \and
Pablo {De Cristóforis}\inst{1,2}\orcidID{0000-0002-7551-7720}}
\authorrunning{F. Raverta Capua et al.}

\institute{Instituto en Ciencias de la Computación (UBA-CONICET), Ciudad Autónoma de Buenos Aires C1428EGA, Argentina \\ \email{fraverta@icc.fcen.uba.ar} \and Departamento de Computación, Facultad de Ciencias Exactas y Naturales, Universidad de Buenos Aires, Ciudad Autónoma de Buenos Aires C1428EGA, Argentina \\ \email{\{jschandin,pdecris\}@dc.uba.ar}} 
%\and
%Springer Heidelberg, Tiergartenstr. 17, 69121 Heidelberg, Germany
\maketitle              % typeset the header of the contribution
\begin{abstract}
%The use of sensor monitoring and forestry robotics has been increasing in recent years, and with it, the used of applied machine learning algorithms. 
Remote sensing through unmanned aerial systems (UAS) has been increasing in forestry in recent years, along with using machine learning for data processing. Deep learning architectures, extensively applied in natural language and image processing, have recently been extended to the point cloud domain. However, the availability of point cloud datasets for training and testing remains limited. 
Creating forested environment point cloud datasets is expensive, requires high-precision sensors, and is time-consuming as manual point classification is required. 
Moreover, forest areas could be inaccessible or dangerous for humans, further complicating data collection. Then, a question arises whether it is possible to use synthetic data to train deep learning networks without the need to rely on large volumes of real forest data.
To answer this question, we developed a realistic simulator that procedurally generates synthetic forest scenes.  Thanks to this, we have conducted a comparative study of different state-of-the-art point-based deep learning networks for forest segmentation. Using created datasets, we determined the feasibility of using synthetic data to train deep learning networks to classify point clouds from real forest datasets. Both the simulator and the datasets are released as part of this work. 
%The abstract should briefly summarize the contents of the paper in

\end{abstract}

\input{intro}

\input{relatedwork}

\input{materialsAndMethods}
\input{results}
\input{conclusions}
\end{document}

%% file: intro.tex
\section{Introduction}
The use of remote sensing for environmental monitoring has grown significantly in recent years, thanks to the development of Terrestrial Laser Scanning (TLS), Aerial Laser Scanning (ALS), and Aerial Photogrammetry, techniques widely used in precision forestry~\cite{ref_articleMurtiyoso2024}. Both LiDAR and camera sensors made it possible to easily acquire three-dimensional data of the studied environment, accurately representing it with high precision level point clouds. Both have been widely used in forest environments for health monitoring, species classification, tree parameter estimation, and even illegal logging detection amidst other applications~\cite{ref_articleGuimarães2020}. Laser scanning, while more expensive, heavier, and energy-consuming, is considered the most accurate method for estimating the forest structure, as both the canopy and the ground can be detected~\cite{ref_articlePessacg2022}.

Deep learning architectures, popular in natural language and image processing nowadays, have recently been extended to point cloud processing, and multiple techniques have already been adapted to the goals of classification, segmentation, and point completion~\cite{ref_articleGuo2020}.%including residual networks \cite{ref_procHe2016}, U-Net networks \cite{ref_procRonneberger2015}, fully convolutional networks \cite{ref_procLong2015} and even attention layers \cite{ref_articleVaswani2017}. 
%\textcolor{blue}{Por otro lado, el uso de técnicas de deep learning se ha extendido a aplicaciones en nubes de puntos. Para esto se han extrapolado muchas técnicas ya utilizadas en otros campos, como multilayered perceptrons [], redes residuales, redes convolucionales, redes de tipo u-net, e incluso transformers, arquitectura muy estudiada hoy en día por sus resultados tan significativos en NLP e imágenes, a clasificación, segmentación, e incluso expansión/refill de nubes de puntos. }

%However, the availability of point cloud datasets for training and testing deep learning architectures remains limited, especially in forest environment 
Unfortunately, few point cloud datasets are publicly available for training, validating, and testing deep learning architectures, among which we can mention ScanNet~\cite{ref_procDai2017}, ScanObjectNN~\cite{ref_procUy2019} and ModelNet40~\cite{ref_procWu2015} for object classification and ShapeNetPart~\cite{ref_articleYi2016} and SemanticKITTI~\cite{ref_procBehley2019} for part segmentation and scene completion. Generally speaking, none of them are designed for specific environments, such as forests. Therefore, if training a deep learning architecture for forested environment is needed, a new dataset for this purpose has to be generated. For example, \cite{ref_articleJin2020} developed a dataset from the regions of the Southern Sierra Nevada Mountains, USA, \cite{ref_articleKrisanski2021} from Australia and New Zealand, and \cite{ref_articleKaijaluoto2022} from Evo, Finland. All these works were conducted for forest segmentation. Of the three, only the latter dataset is publicly available, limiting the repeatability of the experiments and the comparison between the cited works.

%\textcolor{blue}{Hoy en día hay pocos datasets de nubes de puntos disponibles que permitan entrenar, testear y validar nuevas arquitecturas. Entre ellos, los más conocidos son (enumerar). Los objetivos de éstos son en general reconocer objetos, segmentar la escena, y clasificar partes del objeto, respectivamente. En general ninguno de éstos tiene información de ambientes específicos como es el forestal, por lo que si se pretende entrenar alguna arquitectura para poder identificar algún parámetro de interés en este ambiente es necesario generar un dataset manualmente con el que entrenar estos dataset. A modo de ejemplo, (paper de los finlandeses) levantaron un dataset y clasificaron los puntos en las categorías ...., y entrenaron redes para poder segmentar nuevas nubes de puntos tomadas de ambientes forestales en estas categorías. De forma similar, (listar algunos trabajos de los chinos). }

Creating a specific point cloud dataset for forested environments is expensive, as high-end equipment, including UAVs and high-precision sensors, is required to survey the studied area. It is also time-consuming, as it implies labeling the points manually. Moreover, forest areas could be inaccessible or dangerous for humans, further complicating data collection. 

This work aims to answer whether synthetic data for training point-based deep learning networks is suitable for segmenting real forest point clouds generated from LiDAR or camera sensors. For this purpose, we developed a forest simulator based on Unity \cite{ref_urlUnity} that allows us to generate several forested scenes with high realism procedurally. We extract the point clouds from the synthetic scenes, which are then used to train the deep learning architectures instead of using real forest data.  The main contributions of this work are as follows:
\begin{enumerate}
    \item The development of a novel open-source forest simulator based on Unity that procedurally generates forest scenes. It also includes a configurable survey mission planner mode that takes pictures of the scene like a camera from an up-down UAV view. 
    \item Public-domain synthetic datasets of forest scenes that can be used to train or test different deep-learning networks. 
    \item A comparative study of state-of-the-art point-based deep learning networks to determine whether training with synthetic data is suitable for segmenting point clouds from real forest datasets.  
\end{enumerate}

This paper is organized as follows: Section 2 overviews related works. Section 3 briefly explains the deep-learning architectures selected for this work, and presents the developed forest simulator and the dataset generation. Section 4 shows and discusses the experimental results, and Section 5 ends with conclusions and future work.

%% file: relatedwork.tex
\section{Related Work}

The development of TLS, ALS and aerial photogrammetry, aided with computer vision techniques, has made acquiring an accurate 3D reconstruction of the studied environment possible. This allows us to extract valuable information for the characterization, monitoring, and conservation of forests. %Using tools and techniques of computer vision and artificial intelligence models, a digital terrain model (DTM) can be estimated from this reconstruction. In a similar way, the application of aerial photogrammetry and computer vision techniques, such as Structure from Motion, made it possible to compute a 3D model (i.e. point cloud) based on the captured images of the environment, typically taken from a UAV. %These 3D models allow us to extract valuable information for the characterization, monitoring, and conservation of forests, such as structural parameters, tree density, and even identifying areas with excessive or illegal logging~\cite{ref_articleGuimarães2020}.  

Thanks to the ability of the deep learning networks to adapt to several tasks, such as classification, segmentation, and completion, its use has expanded to several applications, including the study of different natural environments. For example,~\cite{ref_articleGevaert2018,ref_articleAmirkolaee2022} use deep learning techniques for the extraction of digital terrain models (DTM) from camera images and~\cite{ref_articleLi2020} uses them for landform classification. Specifically in point cloud processing,~\cite{ref_articleJin2020,ref_articleHu2016,ref_articleLe2022,ref_articleLi2022} use deep learning for the extraction of DTM from LiDAR captured data. In a forest environment,~\cite{ref_articleKrisanski2021} segments the LiDAR captured point cloud into different categories: terrain, vegetation, coarse woody debris (CWD) and stems. It is noticeable that segmenting the terrain from the rest of the points results equivalent to finding the DTM of the studied environment. Similarly,~\cite{ref_articleKaijaluoto2022} uses segmentation techniques to classify the points into terrain, understorey, tree trunk, or foliage categories; the labeled point cloud dataset used in this work is one of the few publicly available. %\cite{ref_articleMa2023} proposes a new architecture specialized in forest environments to segment point clouds into the ground, tree trunks, leaves, and bushes. 
% \textcolor{blue}{En el estado del arte está muy estudiado el uso de técnicas morfológicas para la estimación del modelo de terreno [1], [2], [3]. Sin embargo, estos últimos años crecieron en relevancia los modelos de inteligencia artificial, impulsados por el desarrollo de nuevas tecnologías que permiten generar arquitecturas de redes más precisas con menores tiempos de entrenamiento. En particular, el desarrollo de la tecnología de Transformers [4], muy popular actualmente por su uso en aplicaciones tales como ChatGPT y GPT-4 de OpenAI, que han generado una revolución en el área de Procesamiento Natural del Lenguaje (del inglés Natural Language Processing), y se está buscando trasladar estos mismos resultados a otras áreas de estudio, tales como el Procesamiento Digital de Imágenes [5] y el Procesamiento de Nubes de Puntos (del inglés Point Cloud Processing). }

The development of deep learning architectures designed for point cloud processing started to receive more attention with PointNet++~\cite{ref_articleQi2017}. This architecture was taken as a starting point and as a comparison model for several new networks that seek to classify or segment point clouds. PointNet++ is based on a U-Net architecture~\cite{ref_procRonneberger2015} built mostly with multi-layer perceptrons with an encoder-decoder structure, where the encoder extracts features from point clouds, and the decoder reinterprets the extracted features. Several networks have been based on PointNet++, such as KPConv~\cite{ref_procThomas2019}, PointConv~\cite{ref_procWu2019}, ConvPoint~\cite{ref_articleBoulch}, which incorporated their own convolutional operator designed to work with point clouds, and PFCN~\cite{ref_articleLu2019}, which was implemented using a fully convolutional network. PointNeXt~\cite{ref_articleQian2022} updates PointNet++ to the state-of-the-art using more modern training strategies, such as newer optimization techniques, data augmentation, and modifications of the architecture.% with the incorporation of newer residual connections \cite{ref_procHe2016}, an inverse bottleneck design \cite{ref_procSandler2018} and separable multi-layer perceptrons \cite{ref_articleQian2021}.

% \textcolor{blue}{El desarrollo de arquitecturas pensadas para el Procesamiento de Nubes de Puntos tomó mayor relevancia a partir de PointNet++ [6]. Esta arquitectura se tomó como base y como punto de comparación para numerosas nuevas redes que buscan tanto clasificar como segmentar nubes de puntos. PointNet++ está basada en una arquitectura U-Net [7] del tipo encoder-decoder, el primero para extraer características de las nubes de puntos y el segundo para reinterpretar las características extraídas. Ésta es una arquitectura mayoritariamente basada en multilayer perceptrons.} \textcolor{red}{Agregar acá otras redes más, como la PointConv, ConvPoint, etc.} \textcolor{blue}{PointNeXt [8] es una arquitectura más reciente que retoma PointNet++, y la actualiza al estado del arte al mejorar estrategias de entrenamiento, tales como el uso de técnicas de optimización más recientes o de data augmentation, y al modificar levemente la arquitectura incorporando conexiones residuales nuevas [9], un diseño de inverse bottleneck design [10], y multilayer perceptrons separables [11].}

Transformer technology, based entirely on self-attention layers~\cite{ref_articleVaswani2017}, presents advantages over the more standardized use of convolutional layers and multi-layer perceptrons in encoder-decoder architectures. This translates into more accurate results with lower training time as they are more parallelizable but at the cost of using a greater number of parameters, which requires a greater volume of input data for training.
% \textcolor{blue}{Por otra parte, los Transformers, basados enteramente en capas de self-attention [4] presentan algunas ventajas frente al uso más estandarizado de capas convolucionales y de perceptrones multi-capa (del inglés multilayer perceptrons) en arquitecturas del tipo encoder-decoder. Esto se traduce en resultados más precisos con menores costos de entrenamiento al ser más paralelizables, pero a costa de utilizar una mayor cantidad de parámetros, lo que se traduce en una necesidad de un volumen mayor de datos de entrada.}
The first networks that incorporated transformer technology for point cloud processing were Engel's PointTransformer~\cite{ref_articleEngel2021}, Zhao's PointTransformer~\cite{ref_procZhao2021}, Fast Point Transformer~\cite{ref_procPark2022} and PCT~\cite{ref_articleGuo2021}, all of which try to adapt the structure of transformers into the point cloud data type. Following the same idea, PointBERT~\cite{ref_procYu2022}, that follows the ideas of BERT~\cite{ref_articleDevlin2016} and BEiT~\cite{ref_procBao2022} for natural language processing and image processing respectively, divides the point cloud into several smaller local clouds, and codifies each of them into tokens through a Discrete Variational AutoEncoder~\cite{ref_procRolfe2016}. It then masks a given proportion of the local clouds and uses them as an input to the transformer block built with attention encoders only, which is trained to recover the original tokens of the masked clouds. Then it recovers the clouds from the tokens via a Decoder, and reconstructs the full point cloud.
%\textcolor{red}{Hay que chequear si esto es cierto, y escribir porqué. En el paper dice "we adopted the standard Transformer" y citan a "attention is all you neeed". Hay que chequear lo mismo para MAE y GPT. -> GPT usa decoder, pero reversionado, y según "Transformers in 3D Point Clouds: A Survey", BERT usa el encoder nomás}

% \textcolor{red}{Agregar acá cuáles fueron las primeras con transformers, como PointTransformer, PTF, etc.} \textcolor{blue}{Una de las opciones con más relevancia dentro de Procesamiento de Nubes de Puntos que incorpora el uso de Transformers es PointBERT [12]. Esta arquitectura, que sigue las ideas de BERT [13] y BEiT [14] para Procesamiento Natural del Lenguaje y de Procesamiento de Imágenes respectivamente, divide la nube de puntos en muchas nubes locales, y a cada una de ellas las codifica en tokens mediante un AutoEncoder Variacional Discreto (del inglés Discrete Variational AutoEncoder) [15]. Luego se enmascara cierta proporción de esas nubes locales y con ellas se alimenta el bloque de Transformers, el cual es entrenado para poder recuperar los tokens originales de las nubes enmascaradas. Posteriormente, se recuperan las nubes de puntos a partir de estos tokens mediante un Decoder, reconstruyéndose así la nube de puntos completa. }

Along the same lines, PointMAE~\cite{ref_procPang2022} follows the ideas of MAE~\cite{ref_procHe2022} for Image Processing, dividing the point cloud into smaller local clouds which are masked without the need for a Discrete Variational AutoEncoder. PointGPT~\cite{ref_procChen2023}, inspired by GPT~\cite{ref_articleRadford2018}, also divides the point cloud into smaller local clouds but orders them using a Morton curve by spatial proximity~\cite{ref_articleMorton1966}. It then masks a given proportion of the local clouds with a dual mask strategy and uses them as an input to the transformer block built exclusively with attention decoders. It aims to predict future representations of local clouds in an autoregressive way.

Regarding the use of synthetic datasets in forest environments, in~\cite{ref_procNunes2022}, a simulator is presented using procedural techniques, where sensors like LiDAR, an RGB camera, and a depth camera are also simulated for data extraction, all of them with the capacity of segmenting the scene in the following categories: background, terrain, traversable, trunks, canopy, shrubs, herbaceous plants, and rocks. A synthetic dataset using this simulator is available in~\cite{ref_urlSynPhoRest}. It includes RGB images, semantic segmentation maps, depth maps and the projection of LiDAR point clouds on the RGB field of view for two different LiDAR scanning patterns. However, this simulator is not publicly available, and the dataset does not include the 3D reconstruction neither the point clouds dataset of the environment. Finally,~\cite{ref_procRussell2022} uses this dataset's images for training networks to detect fuel for preventing spread of forest fires. This work concludes that the synthetic data fails to generalize to real data.

%% file: materialsAndMethods.tex
\section{Materials and Methods}

\subsection{Point cloud deep learning networks}

Following \cite{ref_articleKaijaluoto2022}, we aim to train different deep learning networks to segment the forest point clouds into trunks, canopy, understorey, and terrain. This lets us differentiate forest strata and the DTM that corresponds with the terrain points. We selected four well-known state-of-the-art architectures, PointNeXt, PointBERT, PointMAP, and PointGPT, and trained them with synthetic data generated by our forest simulator. Of these four selected architectures, the last three are built using transformers, and pre-trained versions are available with the ShapeNetPart dataset, so these networks can be fine-tuned specializing them in the respective study area. On the other hand, PointNext is not built using transformers but multi-layer perceptrons, and as we do not count on pre-trained versions of the network, we train it from scratch. 

%\textcolor{blue}{Siguiendo la línea de \textcolor{red}{cita a los finlandeses}, queremos entrenar distintas arquitecturas para poder diferenciar distintas clases dentro del ambiente forestal: troncos, follaje, understorey y terreno. De esta forma no sólo estamos encontrando implícitamente el DTM al segmental los puntos correspondientes con el terreno del resto de puntos, sino que además podemos diferenciar distintos estratos arbóreos. Para ésto, elegimos 4 arquitecturas a las que entrenaremos con datos sintéticos de ambientes forestales, guiados por el buen desempeño de éstas: PointNeXt, PointBERT, PointMAP y PointGPT. De éstas, las últimas 3 están construidas utilizando transformers, y están disponibles versiones preentrenadas con el dataset \textcolor{red}{referencia al dataset con el que están entrenadas}, por lo que la red ya cuenta con cierto aprendizaje para la detercción de características realizada con grandes volúmenes de datos. A ésto basta con realizarle una etapa de fine-tuning, que permite especializar la red ya pre-entrenada en el área de estudio respectiva, tal como el ambiente forestal en nuestro caso. Por otra parte, PointNeXt no utiliza transformers, sino que se basa en el uso de multi-layer perceptrons, y dado que no contamos con una versión previamente preentrenada, directamente realizamos el entrenamiento de la red de cero.} 

\subsection{Forest Simulator}

For this work, a forest simulator based on the Unity engine was developed, from which synthetic data with a similar appearance to real-world forests was extracted. We aim to train the mentioned architectures with synthetic data and test their performance with real forest data. %This would save a considerable amount of time and reduce the use of resources necessary to extract in situ information from the places to be explored, classify or label them and use them for training the networks each time a forest study is required. 
As \cite{ref_articleKaijaluoto2022} notices, the manual labeling of the extracted point cloud is a very demanding task, and in several cases, it is impossible to human experts to discern which category each point belongs to. Using a simulator for dataset generation overcomes this problem, as the point cloud labeling can be carried out automatically. Moreover, as transformer technology requires a large volume of data for training, generating synthetic data procedurally becomes even more relevant. Below we detail the most important modules of the presented simulator.

\subsubsection{Terrain generation} The simulator first generates a terrain mesh using fractal noise to build a heightmap for all its vertices. These noise samples are from Perlin noise layers at different scales or octaves. This permits controlling the amount of detail and the general aspect of the terrain. 
As the implementation is easily parallelizable, we took advantage of the Perlin noise function provided by Unity and the Unity Jobs framework for parallel execution. The random appearance and realistic aspect of this method's results are notorious and well-regarded in the video game developers community.

% \textcolor{blue}{El simulador permite generar escenas forestales de forma simple. En primer lugar se genera la malla del terreno utilizando fractal noise, el cual samplea a Perlin Noise Layer a diferentes frecuencias y amplitudes para generar un heightmap para all the vertices of the terrain mesh.} 
\subsubsection{Trees, bushes, and plants generation} Vegetation (excluding grass) is generated via pipelines. Each pipeline is a Directed Acyclic Graph (DAG) that links prefabs (i.e. reusable pre-generated game objects, like individual trees or bushes) with their position over the terrain. This is done through textures that determine the spawn probability and density over the terrain mesh. Each pipeline is built using different nodes:
\begin{itemize}
    \item Source: imports a texture from a file or another pipeline, or generates a new one via a Voronoi diagram or sampling noise.
    \item Logic: applies logic operations over textures
    \item Sampling: variants of Poisson disk sampling method.   
    \item Placement: generates instancing parameters for the assigned prefabs.
\end{itemize}

Regarding the sampling process, an implementation of Bridson's Poisson disk sampling method \cite{ref_articleBridson2007} and a variation of it were implemented. In  Bridson's original algorithm, given an object $a_0$ with radius $r$, new objects of the same radius are added in an annulus of size $[r, 2r]$ without overlapping, until it reaches a maximum quantity, or until there can not be placed any more, and then this process is repeated with the next object. Using a cell size of $r / \sqrt{n}$, where $n$ is the dimension of the background grid for storing samples, each cell can contain only one placed object. This process is fast for object placement, but it produces a distribution of points that may appear equidistant, especially for small values of $r$ giving an unrealistic point distribution. To face this issue we propose a variation to the method: new points are seeked in an annulus of size $[r_{min}, r_{max}]$, where $r_{min}$ and $r_{max}$ are a given minimum radius and a maximum radius respectively, interpolating the distance linearly using the value of a greyscale texture at each point. This means that for values near 0 (where the texture is black), points at a distance $r_{min}$ from $a_0$ are generated, and the inverse holds for values near 1. To do that, we tweaked Bridson's algorithm to use a spatial cell of size $r_{min} / \sqrt{n}$, but instead of each cell holding the index of only one placed object, it holds a list of indices for the objects that shadow that cell and remains at an acceptable distance between themselves.

The texture of spawn probability acts as the probability of effectively instantiating an object at a given point. Having this as a separate node from the sampling process is useful since the number of objects spawned in an area can effectively be reduced, even to 0, making clearings of arbitrary shapes possible.

\begin{figure}[t]
    \begin{subfigure}[b]{0.32\textwidth}
        \includegraphics[width=\textwidth]{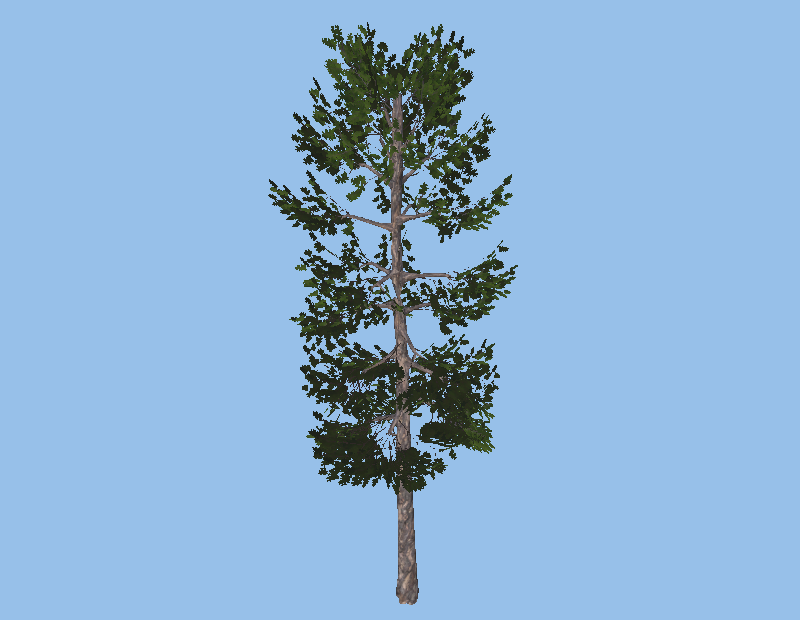}
    \end{subfigure}
    \begin{subfigure}[b]{0.32\textwidth}
        \includegraphics[width=\textwidth]{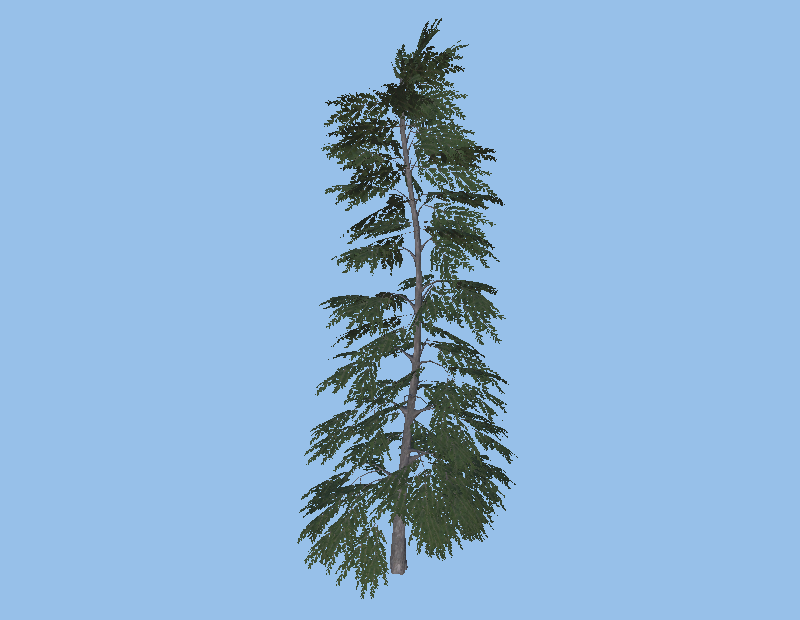}
    \end{subfigure}
    \begin{subfigure}[b]{0.32\textwidth}
        \includegraphics[width=\textwidth]{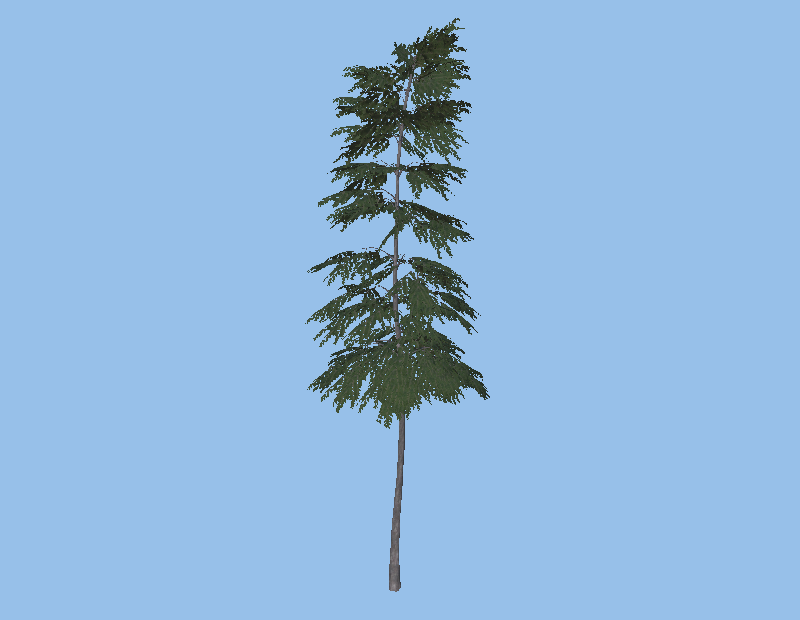}
    \end{subfigure}
    %\begin{subfigure}[b]{0.24\textwidth}
    %    \includegraphics[width=\textwidth]{images/PrefabBush2.png}
    %\end{subfigure}
    %\begin{subfigure}[b]{0.195\textwidth}
    %    \includegraphics[width=\textwidth]{images/PrefabUnderstorey2.png}
    %\end{subfigure}
\caption{Sample models of trees used in the simulator, generated with software TreeIt~\cite{ref_urlTreeIt}.} \label{forestPrefabs}
\end{figure}

There can be created as many pipelines as required. The simulator already counts with basic pipelines for trees and bushes. The prefab models of trees, bushes, and other plants were generated with the free-to-use TreeIt software~\cite{ref_urlTreeIt}. A small sample of them can be seen in Fig.~\ref{forestPrefabs}. More models can be easily added if needed. Before instantiating these models, some transformations are applied to include more variability to the scene: a random spin around its up axis, a random twist to bend the up direction with regards to the world's up direction, and a random scale.

\subsubsection{Grass generation} None of the sampling methods for vegetation distribution could scale to generate millions of points while keeping the frame rate manageable. Thus, a parallelizable method was devised for placing grass. It is generated by indirect instancing, where the geometry is produced via a compute shader (a program that run on the GPU, outside of the normal rendering pipeline) and sent to the graphics pipeline through a shared memory buffer. The sampling algorithm for the grass leaves is depicted in Algorithm~\ref{samplingGrass}. As the tiles are non-overlapping, the processing for each tile can be parallelized over the number of tiles. The generated points are then transformed into the terrain coordinates using parallel raycasting. For a $256 \times 256$ pixels texture, a tile size of 4 pixels and a maximum of 1024 points per tile, the shader takes less than one second to run approximately 4 million points, running in an Intel Core i9-10900 processor, with 32 GB RAM and a NVIDIA GTX 1060 board.

The points are then fed into a compute shader that generates the geometry for a single blade. The number of blade segments can be customized. To add a realistic feeling to the grass, the following transformations are also applied to each blade: a random jitter to the anchor point because of the grid-like pattern of the sampling algorithm; a random rotation that sets which direction the blade is oriented; a random bend for the tip of the blade and a random scaling. After these transformations, the points are returned to world space coordinates to be placed over the terrain. Finally, a shader applies a grass texture to each leaf to add volume and color. 

Approximately 4 million blades of grass, each composed of 9 points, can be generated and updated at $20\sim30$ fps in the mentioned hardware, displaying all grass blades simultaneously. It is worth noting that this instancing method is not used with trees, bushes, and other plants because the random jitter prevents us from enforcing a minimum radius distance between instanced objects to avoid collisions. 

\begin{algorithm}[t]
\caption{Grass Sampling}\label{samplingGrass}
\hspace*{\algorithmicindent} \textbf{Input:} texture $T$, tile size $t_s$, maximum number of points for a tile $P$\\
\hspace*{\algorithmicindent} \textbf{Output:} points for grass leaf instancing $G$
\begin{algorithmic}[1]
\State Divide the texture $T$ into non-overlapping tiles according to a tile size $t_s$
\State Define an empty list $G$ for the points for grass instancing
\For{each tile $t$}
        \State Define density $d = \frac{1}{t_s^2} \sum\limits_{(i,j) \in t} p_{i,j}$, where $p_{i,j}$ is the texture's value in pixel $(i,j)$
        \State Define the number of sampling points in $t$ by $p = d \times P$
        \State Define points $G_t$ by distributing $p$ in a grid-like pattern in the tile
        \State Append $G_t$ to $G$
\EndFor
\State \Return generated points $G$
\end{algorithmic}
\end{algorithm}

\subsubsection{Repeatability} 
\begin{figure}[h!]
    \centering
    %\begin{subfigure}[b]{\textwidth}
    %    \includegraphics[width=0.98\textwidth]{images/frontish_view.png}
    %\end{subfigure}
    %\\
    \begin{subfigure}[b]{0.46\textwidth}
        \includegraphics[width=\textwidth, angle = 180]{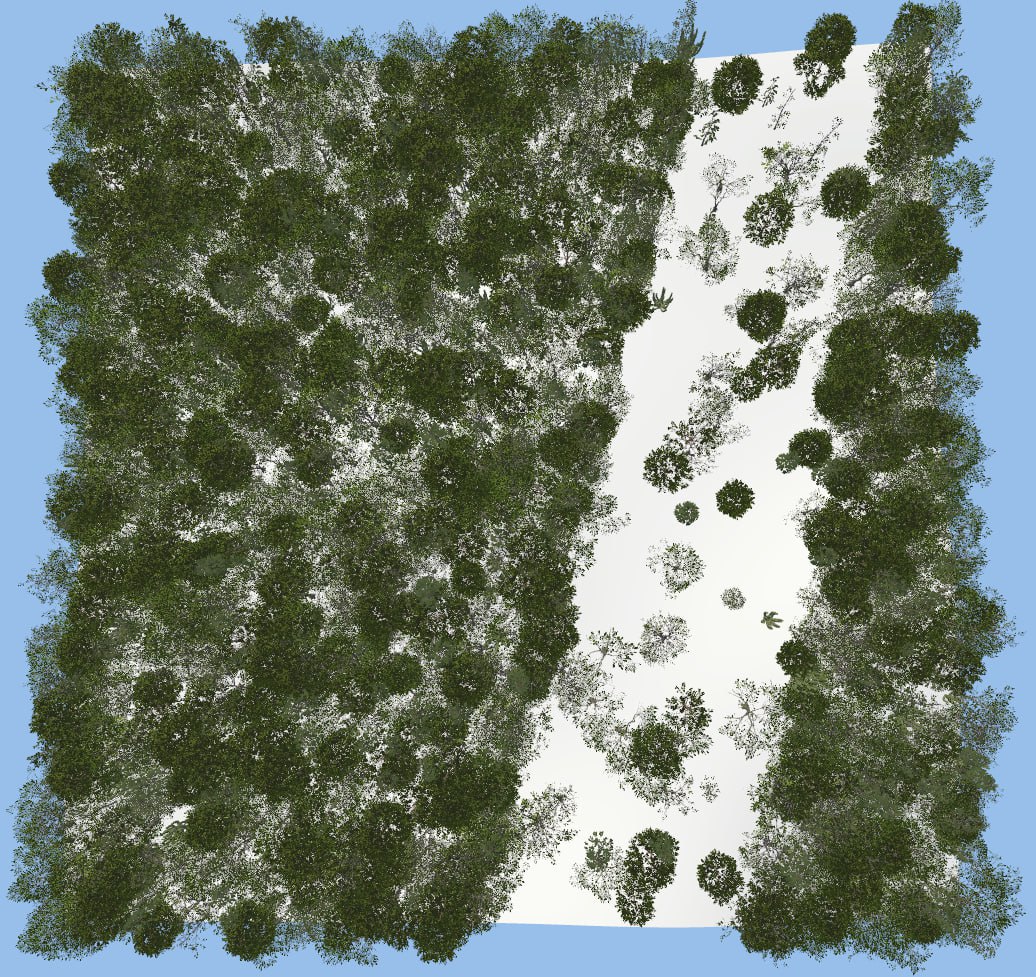}
    \end{subfigure}
    %\begin{subfigure}[b]{0.32\textwidth}
    %    \includegraphics[width=\textwidth]{images/photobushes.jpeg}
    %\end{subfigure}
    \begin{subfigure}[b]{0.46\textwidth}
        \includegraphics[width=\textwidth, angle = 180]{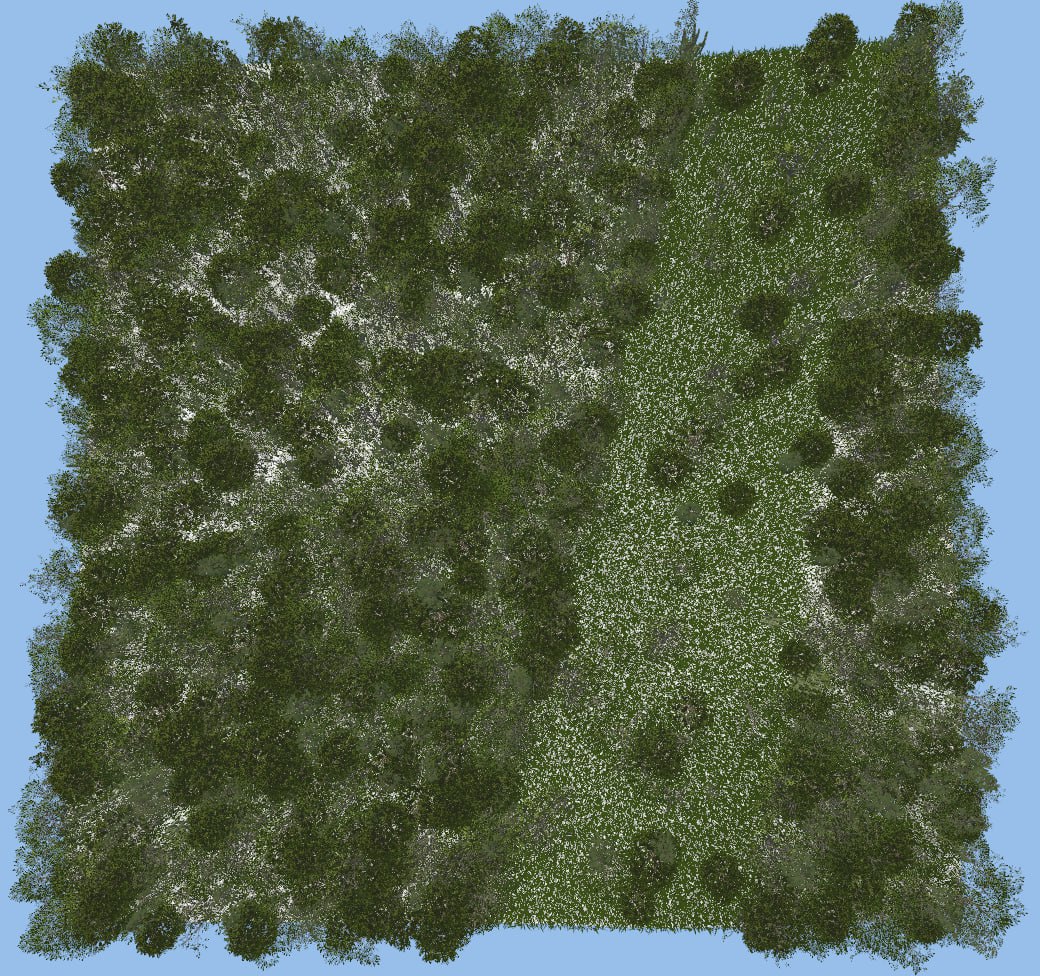}
    \end{subfigure}
    \\
    \begin{subfigure}[b]{0.46\textwidth}
        \includegraphics[width=\textwidth]{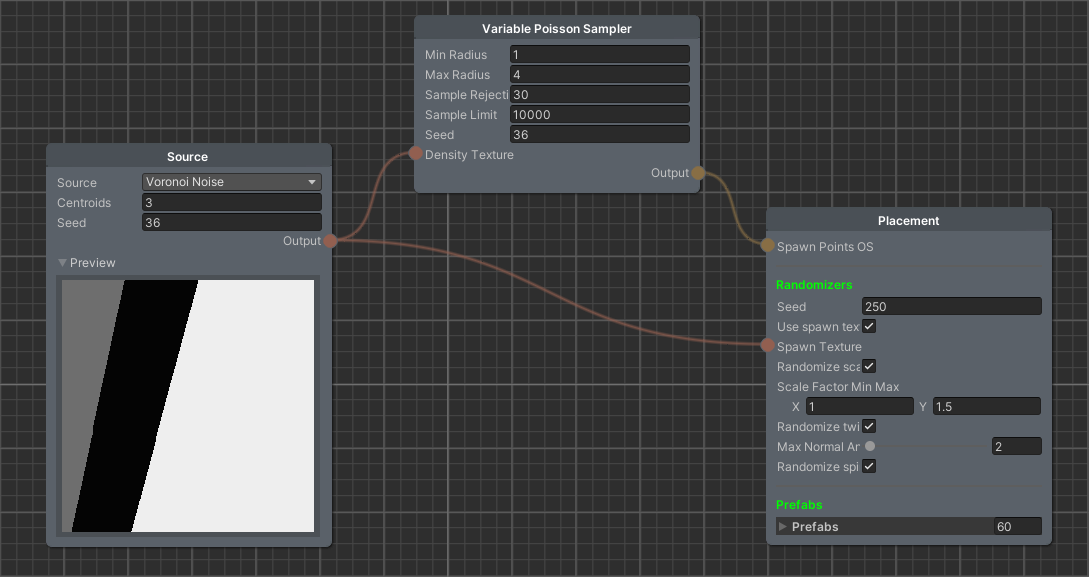}
    \end{subfigure}
    %\begin{subfigure}[b]{0.32\textwidth}
    %    \includegraphics[width=\textwidth]{images/Pipeline_bushes.png}
    %\end{subfigure}
    \begin{subfigure}[b]{0.46\textwidth}
        \includegraphics[width=\textwidth]{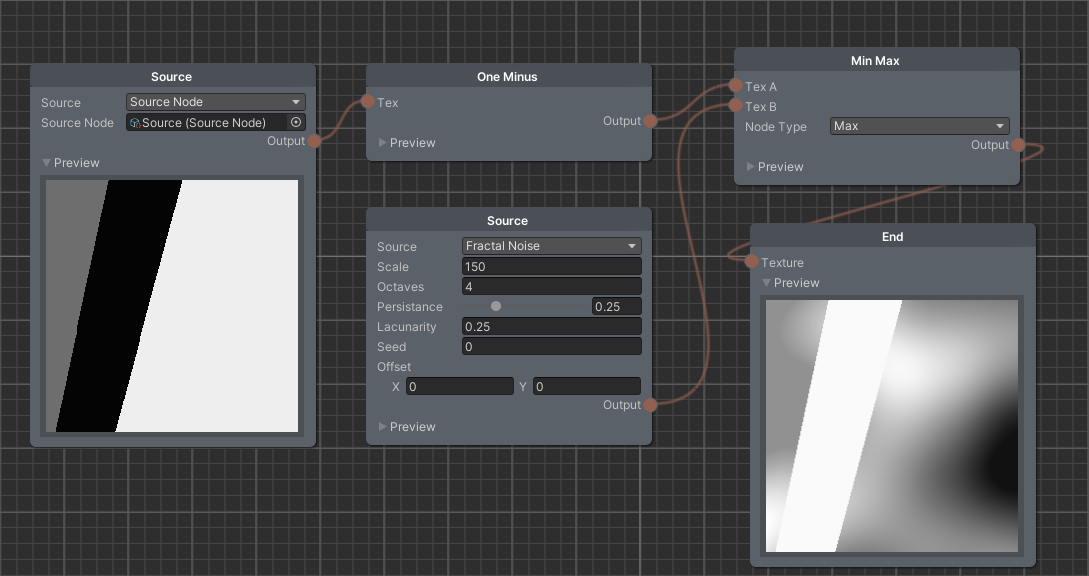}
    \end{subfigure}
\caption{Above: Forest scene with trees (left), and trees, bushes, and grass (right). Below: the correspondence pipelines for instancing trees (left), and grass (right).} \label{forestSceneUnity}
\end{figure}

\begin{figure}[h!]
    \centering
    \begin{subfigure}[b]{0.506\textwidth}
        \includegraphics[width=\textwidth]{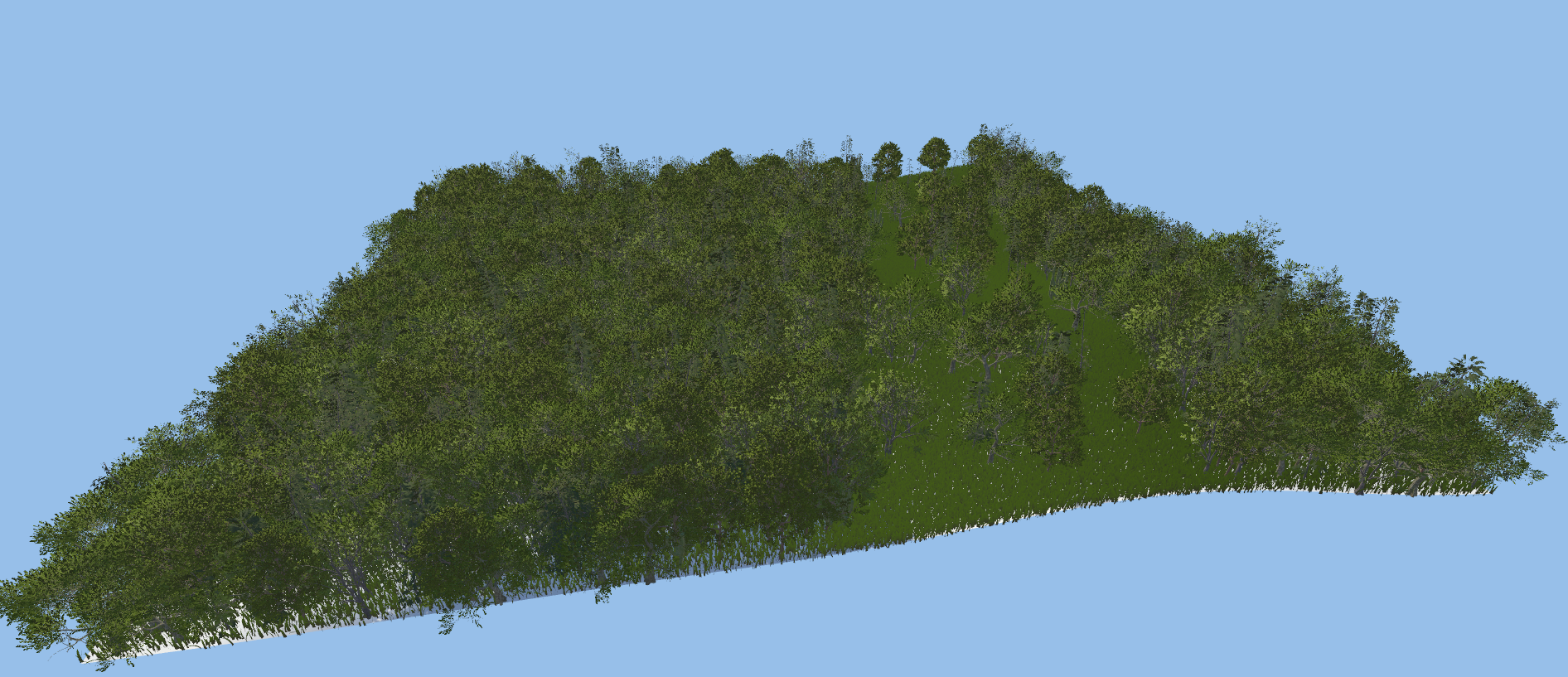}
    \end{subfigure}
    \begin{subfigure}[b]{0.414\textwidth}
        \includegraphics[width=\textwidth]{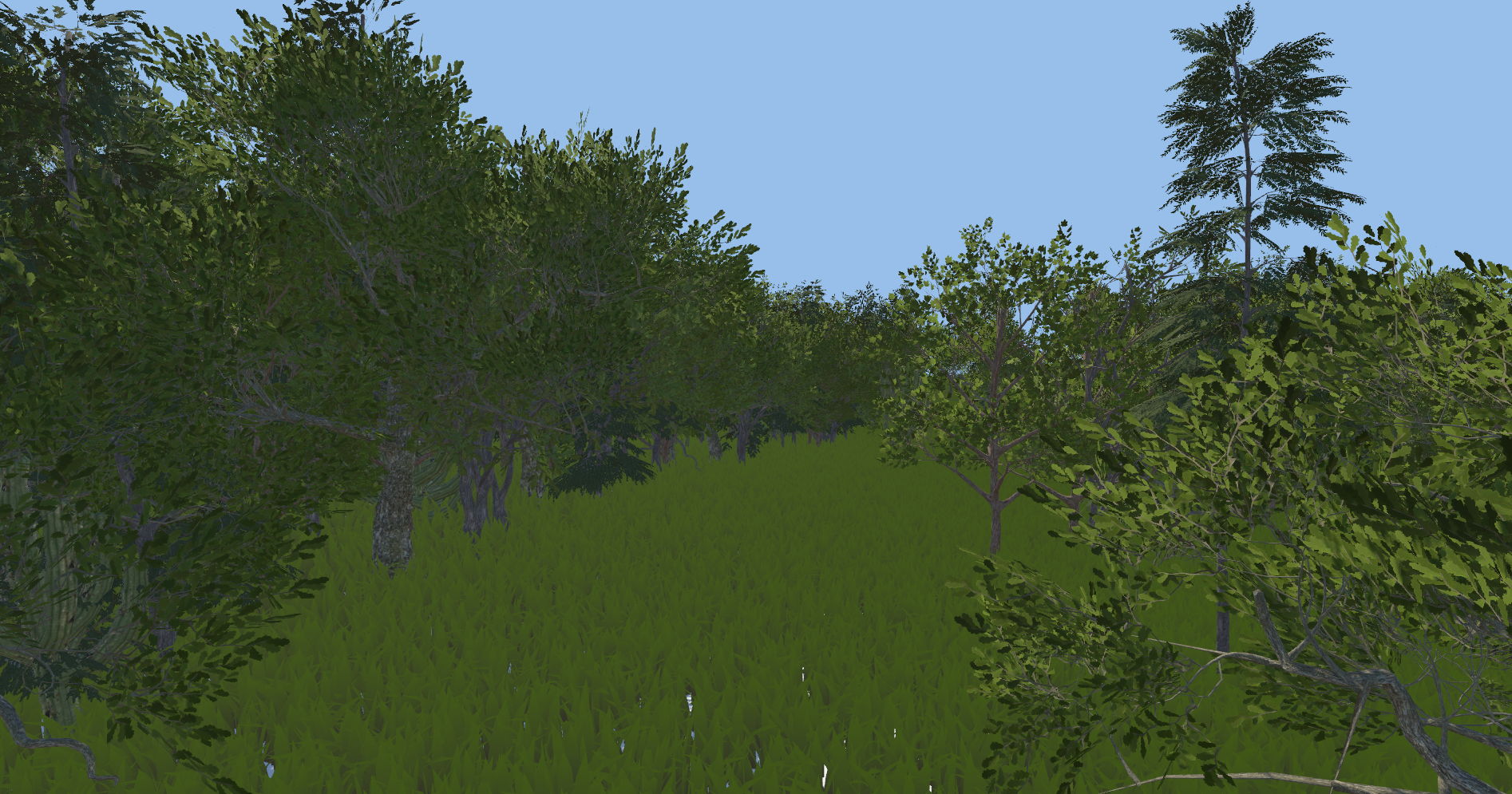}
    \end{subfigure}
\caption{Left: Frontal view of a generated forest scene. Right: Close up view.} \label{forestSceneUnityClose}
\end{figure}

Each scene is generated by a seed to ensure repeatability. This seed is transmitted to every vegetation pipeline and to the grass and terrain generators. Fig.~\ref{forestSceneUnity} shows an example of the pipelines for generating trees and grass and a top-down view of the resulting forest scene, and Fig.~\ref{forestSceneUnityClose} shows the same scene from a front view and a closer view. 

%\textcolor{blue}{La generación de cada escena está vinculada con una semilla, lo que permite la repetitibilidad de cada escena formada. Ésta semilla se transmite a todos los pipelines, y a la generación del cesped/pastizal.} 

\subsubsection{Point Cloud Extraction} The point cloud of the generated scene can be extracted directly from the Unity Editor as a .csv file. By tagging the instanced objects with meshes, they can be exported as various categories, including but not limited to terrain, canopy, trunk, branches, bushes, understorey, grass, cacti, and deadwood, assigning the corresponding label to each point of the point cloud. The size of the scene's point cloud can be altered by adding more points to the terrain mesh, by generating more grass leaves or changing their number of segments, or by importing other vegetation prefabs with the desired quantity. This customization helps generating scenes where its point cloud can vary in size, and thus be adapted to specific needs, such as training large deep learning networks. 

As one of the contributions of this work, the code of the presented forest simulator was released at \url{https://github.com/lrse/forest-simulator}.

%\textcolor{blue}{Para el desarrollo del simulador se utilizaron modelos de árboles, arbustos, plantas y pastizales generados con el software TreeIt [referencia], de uso libre y gratuito.} \textcolor{red}{Poner alguna imagen de cómo queda una escena instanciada, que sea la misma que la de los mapas de densidad puestos como ejemplo. }

%\textcolor{blue}{La extracción de las nubes de puntos que conforman las mallas se realiza desde el mismo editor de Unity}.
%\textcolor{blue}{, seleccionando de la escena qué elementos se desea extraer de forma individual. Para esto se utilizó la herramienta [], provista en el Asset Store de Unity. Para este trabajo, y usando como referencia a [], se exportaron las mallas correspondientes al suelo, a los troncos de los árboles, al follaje de los árboles, y al resto de la vegetación por separado.} \textcolor{red}{Poner alguna imagen de las mallas por separado. }

%\textcolor{red}{Poner una imágen del grafo a modo explicativo, y una de la escena resultante donde se vea la relación de densidades entre las texturas y la escena}

\subsection{Dataset Assembling}

Two datasets were created to train the selected deep learning architectures, simulating the point clouds obtained by LiDAR and applying structure from motion algorithm to synthetic camera images, both from a top-down view. For the camera-like dataset, a method to include occlusion to the point cloud was used~\cite{ref_articleKatz2007}, as several points should not be visible from a top-down view of the forest. Then, random noise with zero mean is added to give variability to the point clouds, and it is partitioned in the $xy$ plane using $K$-Means clustering to assemble subclouds with which to train the networks. Both datasets, LiDAR-like and Camera-like, are publicly at \url{https://github.com/lrse/synthetic-forest-datasets} to ensure this work's results are reproducible and to increase the point cloud datasets available to the scientific community.

%\textcolor{blue}{Para entrenar las redes seleccionadas se realizaron dos datasets distintos, imitando las nubes de puntos generadas por un LiDAR y por una cámara, respectivamente. Para este último se agregó oclusión desde un ángulo zenital, de forma de filtrar los puntos que resultarían no visibles. La estructura general puede verse en el pseudocódigo []. Ambos datasets se encuentran disponibles en [].}

%\textcolor{red}{La función de oclusión usa: Katz et al. ‘Direct Visibility of Point Sets’, 2007}

%\begin{itemize}
%    \item \textcolor{blue}{Levantar una nube de puntos uniendo todas las categorías}
%    \item \textcolor{blue}{Realizar la oclusión mediante el algoritmo de [Katz, 'Direct Visibility of Pointsets']}
%    \item \textcolor{blue}{intersectar la nube de puntos filtrada con la de cada categoría, obteniendo las categorías filtradas resultantes}
%    \item \textcolor{blue}{Agregar ruido a las nubes de puntos de cada categoría}
%    \item \textcolor{blue}{Anexar índice por categoría (0, terreno, ...), unirlas, mezclar aleatoriamente}
%    \item \textcolor{blue}{Particionar en subnubes/clusters sobre el plano x, y de 10000 puntos con Kmeans}
%    \item \textcolor{blue}{guardar las nubes de puntos}
%\end{itemize}

The Evo Dataset, given by \cite{ref_articleKaijaluoto2022}, was employed to test the trained architectures. We used an occluded version of this dataset to test the architectures trained with the Camera-like dataset, using the same occlusion method~\cite{ref_articleKatz2007}.

\begin{figure}[h]
    \centering
    \begin{subfigure}[b]{0.42\textwidth}
        \includegraphics[width=\textwidth]{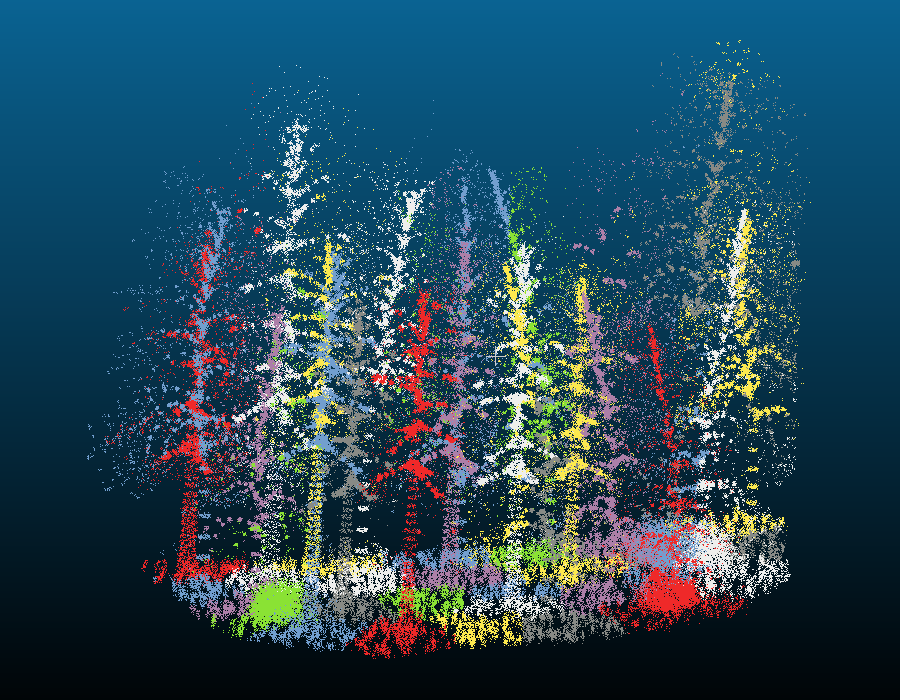}
    \end{subfigure}
    \begin{subfigure}[b]{0.42\textwidth}
        \includegraphics[width=\textwidth]{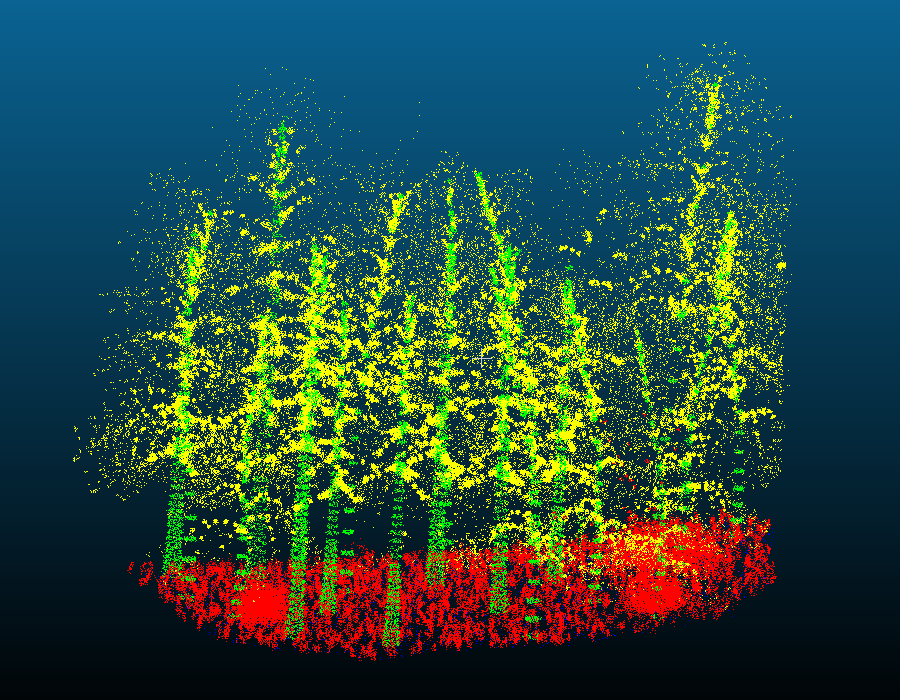}
    \end{subfigure} \\
    \begin{subfigure}[b]{0.42\textwidth}
        \includegraphics[width=\textwidth]{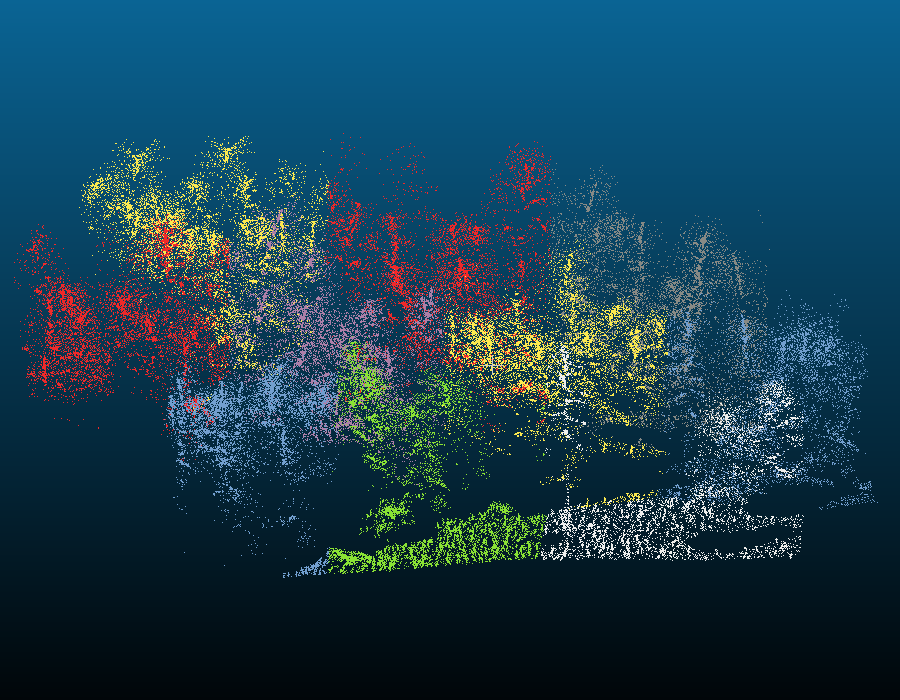}
    \end{subfigure}
    \begin{subfigure}[b]{0.42\textwidth}
        \includegraphics[width=\textwidth]{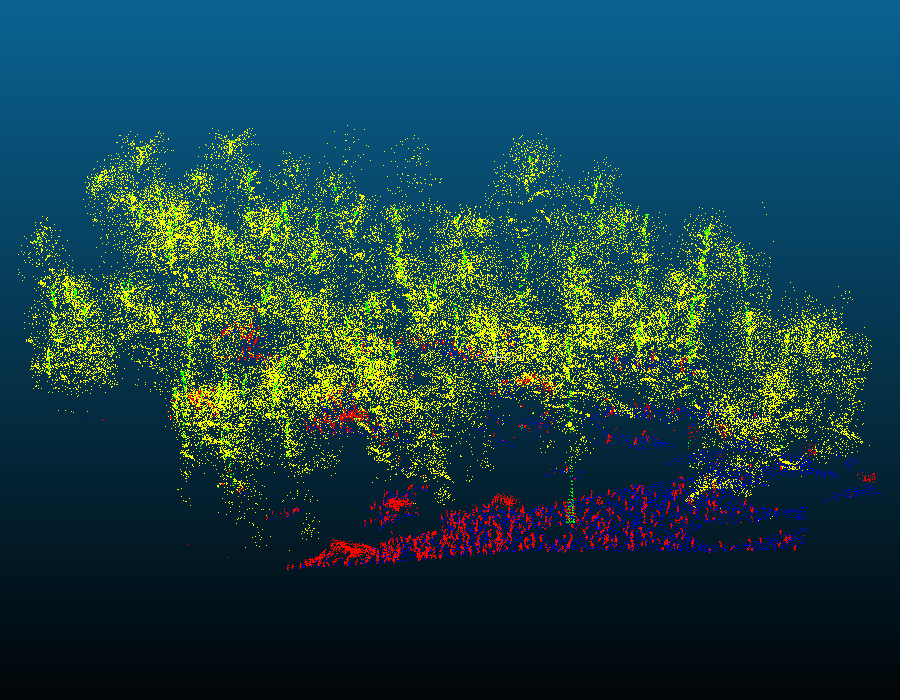}
    \end{subfigure}
\caption{Above: Example scene without occlusion segmented using $K$-Means (left) and segmented via labels: blue, green, yellow, and red corresponding to terrain, trunks, canopy, and understorey, respectively (right). Below: Same example for another scene, but using occlusion from a top-view point.} \label{datasetsPoints}
\end{figure}

%\textcolor{blue}{Para validar el entrenamiento se utilizará el dataset provisto por []. Se le agregará oclusión de análoga a la realizada con el dataset con oclusión para su validación.}
To create a synthetic dataset that resembles the Evo Dataset, we have extracted four categories from the simulator: terrain, trunks, canopy, and understorey, the latter including grass, bushes, and all other vegetation that are not trees. 
In Fig.~\ref{datasetsPoints}, an example of both datasets segmented using $K$-means and segmented into the studied categories can be seen. 

%% file: results.tex
\section{Results and Discussion}

For experimental results, the four selected architectures were trained with an AMD Phenom II X6 1075T Processor CPU, with 32GB RAM, and two NVIDIA RTX 3090 connected with an SLI bridge. 50 epochs were used for every network, with an average training time of 472 seconds per epoch for the LiDAR-like dataset and 239 seconds for the Camera-like dataset.

\subsection{LiDAR-like experiment}

Table~\ref{results_without_occlusion} shows the results of testing the networks trained with the LiDAR-like dataset with the Evo dataset. The confusion matrix of each network is presented in Table~\ref{dispersionMatrices_withoutOcc}. It can be seen that, regardless of having an overall good accuracy, the networks still struggle to differentiate understorey from terrain, especially when the vegetation is on a near-ground level. Despite this, in Section~\ref{section4.3} we  show that the networks perform well in segmenting the trees from the rest of the categories. PointNeXt had a better result in classifying the points in the Evo dataset and segmenting each category, and therefore it seems more suitable for LiDAR datasets. 

\begin{table}[H]
\caption{Results obtained testing both on the created Synthetic Forest Dataset and in the Evo dataset, using the LiDAR-like dataset.}\label{results_without_occlusion}
\resizebox{\linewidth}{!}{
\begin{tabular}{l | c c c|c c c}
\multirow{3}{*}{Network} &
      \multicolumn{3}{c|}{\textbf{Sythetic Forest Dataset}} &
      \multicolumn{3}{c}{\textbf{Evo Dataset}} \\ \cline{2-7}
& \multirow{2}{*}{\shortstack[c]{Overall \\Accuracy}} & \multirow{2}{*}{\shortstack[c]{Class Avg.\\ Accuracy}} & \multirow{2}{*}{\shortstack[c]{Class Avg.\\ mIoU}}  & \multirow{2}{*}{\shortstack[c]{Overall \\ Accuracy}} & \multirow{2}{*}{\shortstack[c]{Class Avg.\\ Accuracy}} & \multirow{2}{*}{\shortstack[c]{Class Avg. \\ mIoU} }  \\ & & & & & & \\
\hline
PointNeXt & 0.9348 & 0.8339 & 0.7314 & \textbf{0.7695} & 0.6436 & \textbf{0.5153} \\
PointBERT & 0.9551 & 0.8719 & 0.8021 & 0.7195 & \textbf{0.6462} & 0.4206  \\
PointMAE & \textbf{0.9575} &\textbf{ 0.8743} & \textbf{0.8029} & 0.72788 & 0.61103 & 0.4278 \\
PointGPT & 0.9414 & 0.8257 & 0.7516 & 0.6417 & 0.5271 & 0.3620   \\
\end{tabular} }
\end{table}

\begin{table}[H]
\caption{Confusion matrix for the four studied networks.}\label{dispersionMatrices_withoutOcc}
\resizebox{.495\linewidth}{!}{
  \begin{tabular}{l|c c c c}
    %\cline{2-5}
    \multicolumn{5}{c}{\textbf{PointNeXt}} \\
     & Terrain & Trunk & Canopy & Understorey \\
    \hline
    Terrain & \textbf{1055469} & 4234 & 423 & 386498 \\
    Trunk & 468 & \textbf{191869} & 135275 & 59160  \\
    Canopy & 10201 & 254302 & \textbf{8310787} & 1220544  \\
    Understorey & 1105198 & 21323 & 1067 & \textbf{1128622} \\
    %\hline
    \end{tabular} }
\resizebox{.495\linewidth}{!}{
  \begin{tabular}{l|c c c c}
    %\cline{2-5}
      \multicolumn{5}{c}{\textbf{PointBERT}}  \\ %\cline{2-5}
    &  Terrain & Trunk & Canopy & Understorey\\
    \hline
    Terrain & 756515 & 16905 & 6451 & \textbf{1314363} \\
    Trunk & 3386 & \textbf{401490} & 184272 & 67868 \\
    Canopy &  11031 & 1088940 & \textbf{6703359} & 765463 \\
    Understorey & 373703 & 38963 & 22493 & \textbf{2129238} \\
    \end{tabular} }
    \newline
    \vspace*{0.1cm} % Adjust the space as needed
    \newline
    
    \resizebox{.495\linewidth}{!}{
  \begin{tabular}{l|c c c c}
      \multicolumn{5}{c}{\textbf{PointMAE}}  \\ 
    & Terrain & Trunk & Canopy & Understorey \\
    \hline
    Terrain & 315040 & 2506 & 18805 & \textbf{1756467}  \\
    Trunk & 193 & \textbf{340098} & 187716 & 129111 \\
    Canopy & 6316 & 779671 & \textbf{6986603} & 796919 \\
    Understorey & 49187 & 13302 & 38252 & \textbf{2465254}  \\

  \end{tabular} }
\resizebox{.495\linewidth}{!}{
  \begin{tabular}{l|c c c c}
      \multicolumn{5}{c}{\textbf{PointGPT}}  \\ 
    & Terrain & Trunk & Canopy & Understorey \\
    \hline
     Terrain & 363932 & 8427 & 13425 & \textbf{1709711} \\
     Trunk & 593 & 219103 & 206613 & \textbf{229875} \\
     Canopy & 11898 & 1049754 & \textbf{5682647} & 1824946 \\
     Understorey & 112227 & 23866 & 31023 & \textbf{2387380} \\
  \end{tabular} }
\end{table}

\subsection{Camera-like experiment}

Table~\ref{results_with_occlusion} shows the results of testing the networks trained with the Camera-like dataset with the Evo dataset with occlusion. The confusion matrices of each network can be seen in Table~\ref{dispersionMatrices_occlusion}. Similar to the previous case, the networks struggle to differentiate terrain points from understorey points, and as few trunk points remain visible, especially the ones at the base of the trunk, it is also often confused with understorey points. We can see that the overall performance is significantly lower than in the case without occlusion. Again, when considering only two categories, Tree and Non-tree, the overall accuracy achieves better results (see Section~\ref{section4.3}). PointMAE obtains slightly better accuracy than the other networks, although all have similar responses.

\begin{table}[H]
\caption{Results obtained testing both on the created Synthetic Forest Dataset and in the Evo dataset, using the Camera-like dataset.}\label{results_with_occlusion}
\resizebox{\linewidth}{!}{
\begin{tabular}{l|c c c| c c c}

\multirow{3}{*}{Network} &
      \multicolumn{3}{c|}{\textbf{Sythetic Forest Dataset}} &
      \multicolumn{3}{c}{\textbf{Evo Dataset}} \\ 
\cline{2-7}
 & \multirow{2}{*}{\shortstack[c]{Overall \\Accuracy}} & \multirow{2}{*}{\shortstack[c]{Class Avg.\\ Accuracy}} & \multirow{2}{*}{\shortstack[c]{Class Avg.\\ mIoU}}  & \multirow{2}{*}{\shortstack[c]{Overall \\ Accuracy}} & \multirow{2}{*}{\shortstack[c]{Class Avg.\\ Accuracy}} & \multirow{2}{*}{\shortstack[c]{Class Avg. \\ mIoU} }  \\ & & & & & & \\
\hline
PointNeXt & \textbf{0.9497} & 0.6875 & 0.5926 &  0.6236 & 0.4868 & 0.4120  \\
PointBERT & 0.9369 & 0.7246 & 0.6303 &  0.6549 & 0.5085 & \textbf{0.4533} \\
PointMAE & 0.9401 & \textbf{0.7289} & \textbf{0.6419} & \textbf{0.6822} & \textbf{0.5448} & 0.4248\\
PointGPT & 0.9336 & 0.7180 & 0.6175 & 0.5794 & 0.5445 & 0.3909 \\

\end{tabular} }
\end{table}

\begin{table}[H]
\caption{Confusion matrix for the four studied networks.}\label{dispersionMatrices_occlusion}
\resizebox{.495\linewidth}{!}{
  \begin{tabular}{l|c c c c}
      \multicolumn{5}{c}{\textbf{PointNeXt}} \\ 
    & Terrain & Trunk & Canopy & Understorey \\
    \hline
    Terrain & \textbf{6566} & 0 & 0 & 6229 \\
    Trunk & 68 & 0 & 24 & \textbf{299} \\
    Canopy & 100 & 0 & \textbf{6990} & 909 \\
    Understorey & 2391 & 0 & 0 & \textbf{3048}  \\
    \end{tabular} }
\resizebox{.495\linewidth}{!}{
  \begin{tabular}{l|c c c c}
      \multicolumn{5}{c}{\textbf{PointBERT}} \\ 
    & Terrain & Trunk & Canopy & Understorey\\
    \hline
    Terrain &\textbf{8923} & 0 & 0 & 5388 \\
    Trunk & 102 & 23 & 40 & \textbf{484} \\
    Canopy & 140 & 2 & \textbf{6226} & 639 \\
    Understorey & \textbf{2392} & 0 & 0 & 2265 \\
    \end{tabular} }
    \newline
    \vspace*{0.2cm} % Adjust the space as needed
    \newline
    \resizebox{.495\linewidth}{!}{
    \begin{tabular}{l|c c c c}
      \multicolumn{5}{c}{\textbf{PointMAE}}  \\
    & Terrain & Trunk & Canopy & Understorey\\
    \hline
    Terrain & \textbf{9508} & 0 & 0 & 4653  \\
    Trunk & 71 & 75 & 38 & \textbf{445}  \\
    Canopy & 108 & 18 & \textbf{6028} & 951  \\
    Understorey & 2176 & 0 & 0 & \textbf{2553}  \\
  \end{tabular} }
      \resizebox{.495\linewidth}{!}{
    \begin{tabular}{l|c c c c}
      \multicolumn{5}{c}{\textbf{PointGPT}} \\ 
    & Terrain & Trunk & Canopy & Understorey\\
    \hline
    Terrain & \textbf{7283} & 3 & 0 & 7069 \\
    Trunk & 80 & 219 & 25 & \textbf{308} \\
    Canopy & 137 & 415 & \textbf{5278} & 1126 \\
    Understorey & 2018 & 17 & 0 & \textbf{2646} \\
  \end{tabular} }

\end{table}

\subsection{Tree and Non-Tree segmentation}
\label{section4.3}
Table~\ref{results_trees_nontrees} shows the results obtained considering only Tree and Non-Tree categories, the first one including trunk and canopy points, and the latter including terrain and understorey points, training with both datasets, LiDAR-like and Camera-like, and testing with the Evo Dataset. It can be seen that the tree segmentation obtains a high accuracy percentage with all the considered networks, being PointBERT the one with better results. 

\begin{table}[t]
\caption{Results obtained considering the categories Tree and Non-Tree, testing in the Evo dataset and training in both the LiDAR-like and the Camera-like datasets.}\label{results_trees_nontrees}
%\caption{Resultados obtenidos considerando únicamente las categorías "Árbol" y "No Árbol", al testear con el Dataset Evo, al entrenar con los datasets LiDAR-like y Camera-like.}
\resizebox{\linewidth}{!}{
\begin{tabular}{l | c c c|c c c}
\multirow{3}{*}{Network} &
      \multicolumn{3}{c|}{\textbf{LiDAR-like Dataset}} &
      \multicolumn{3}{c}{\textbf{Camera-like Dataset}} \\ \cline{2-7}
& \multirow{2}{*}{\shortstack[c]{Overall \\Accuracy}} & \multirow{2}{*}{\shortstack[c]{Class Avg.\\ Accuracy}} & \multirow{2}{*}{\shortstack[c]{Class Avg.\\ mIoU}}  & \multirow{2}{*}{\shortstack[c]{Overall \\ Accuracy}} & \multirow{2}{*}{\shortstack[c]{Class Avg.\\ Accuracy}} & \multirow{2}{*}{\shortstack[c]{Class Avg. \\ mIoU} }  \\ & & & & & & \\
\hline
PointNeXt & 0.9051 & 0.9330 & 0.8035 & 0.9414 & \textbf{0.9180} & 0.8765 \\
PointBERT & \textbf{0.9328} & \textbf{0.9449} & \textbf{0.8652} & \textbf{0.9487} & 0.9108 & \textbf{0.8773}  \\
PointMAE  & 0.9276 & 0.9416 & 0.8561 & 0.9408 & 0.8982 & 0.8596 \\
PointGPT  & 0.8454 & 0.8797 & 0.7252 & 0.9372 & 0.8906 & 0.8497   \\
\end{tabular} }
\end{table}

%% file: conclusions.tex
\section{Conclusions and future work}
In this work, we developed an open-source simulator based on Unity Engine that generates realistic forest scenes procedurally. With it, we have created synthetic point-based datasets, with each point labeled into one of the categories: terrain, trunk, canopy, and understorey (including grass, bushes, and other vegetation that are not trees). We then employed these datasets to train four state-of-the-art deep-learning point-based networks. Finally, we tested and compared them in the real forest Evo dataset. The results show that synthetic point cloud data can be used to train deep-learning networks for posterior forest segmentation with real data. Among the tested networks, PointNeXt seemed to give better results when trained with the LiDAR-like dataset than the other networks, whereas PointMAE obtained slightly better accuracy when trained with the Camera-Like dataset. When considering only two categories, Tree and Non-tree, a higher accuracy percentage is obtained, and PointBERT achieved better results than the other networks. 

As future work, we will use synthetic data to pre-train the deep learning networks and then fine-tune them using real data, expecting better accuracy, specially in the experiments that considers the four categories (terrain, trunks, canopy and understorey). This result is also relevant since it would allow data collection in only a portion of the forest area of interest to conduct deep learning network training.